\documentclass[letterpaper, 10 pt,conference]{IEEEtran}
\IEEEoverridecommandlockouts
\usepackage{cite}
\usepackage{amsmath,amssymb,amsfonts}
\usepackage{algorithmic}
\usepackage{graphicx}
\usepackage{textcomp}
\usepackage{xcolor}
\def\BibTeX{{\rm B\kern-.05em{\sc i\kern-.025em b}\kern-.08em
    T\kern-.1667em\lower.7ex\hbox{E}\kern-.125emX}}
\usepackage{multirow}
\usepackage{subfigure}
\begin{document}

\title{\LARGE \bf
\vspace{0.55cm}
Multimodal Behaviour Trees for Robotic Laboratory Task Automation
}


\author{Hatem Fakhruldeen$^*$$^{1}$, Arvind Raveendran Nambiar$^*$$^{2}$, Satheeshkumar Veeramani$^*$$^{1}$, \\ Bonilkumar Vijaykumar Tailor$^{2}$, Hadi Beyzaee Juneghani$^{1}$, Gabriella Pizzuto$^{2}$ and Andrew Ian Cooper$^{1}$
\thanks{*denotes equal contribution. 
\newline Corresponding author: \textit{h.fakhruldeen@liverpool.ac.uk}
\newline $^{1}$Dept. of Chemistry, University of Liverpool, UK; $^{2}$Dept. of Computer Science, University of Liverpool, UK.
\newline This work was supported by the Leverhulme Trust through the Leverhulme Research Centre for Functional Materials Design, the H2020 ERC Synergy Grant Autonomous Discovery of Advanced Materials under grant agreement no. 856405, the Engineering and Physical Sciences Research Council (EPSRC) under grant agreements EP/V026887/1 and EP/Y028759/1, and the Royal Academy of Engineering under the Research Fellowship Scheme.
}
}


\maketitle


\begin{abstract}
Laboratory robotics offer the capability to conduct experiments with a high degree of precision and reproducibility, with the potential to transform scientific research.
Trivial and repeatable tasks; e.g., sample transportation for analysis and vial capping are well-suited for robots; if done successfully and reliably, chemists could contribute their efforts towards more critical research activities. 
Currently, robots can perform these tasks faster than chemists, but how reliable are they? Improper capping could result in human exposure to toxic chemicals which could be fatal. 
To ensure that robots perform these tasks as accurately as humans, sensory feedback is required to assess the progress of task execution. 
To address this, we propose a novel methodology based on behaviour trees with multimodal perception. 
Along with automating robotic tasks, this methodology also verifies the successful execution of the task, a fundamental requirement in safety-critical environments.
The experimental evaluation was conducted on two lab tasks: sample vial capping and laboratory rack insertion. 
The results show high success rate, i.e., 88\% for capping and 92\% for insertion, along with strong error detection capabilities. 
This ultimately proves the robustness and reliability of our approach and that using multimodal behaviour trees should pave the way towards the next generation of robotic chemists.

\end{abstract}

\begin{figure}[h]
    \centering
    \includegraphics[width=0.3\textwidth]{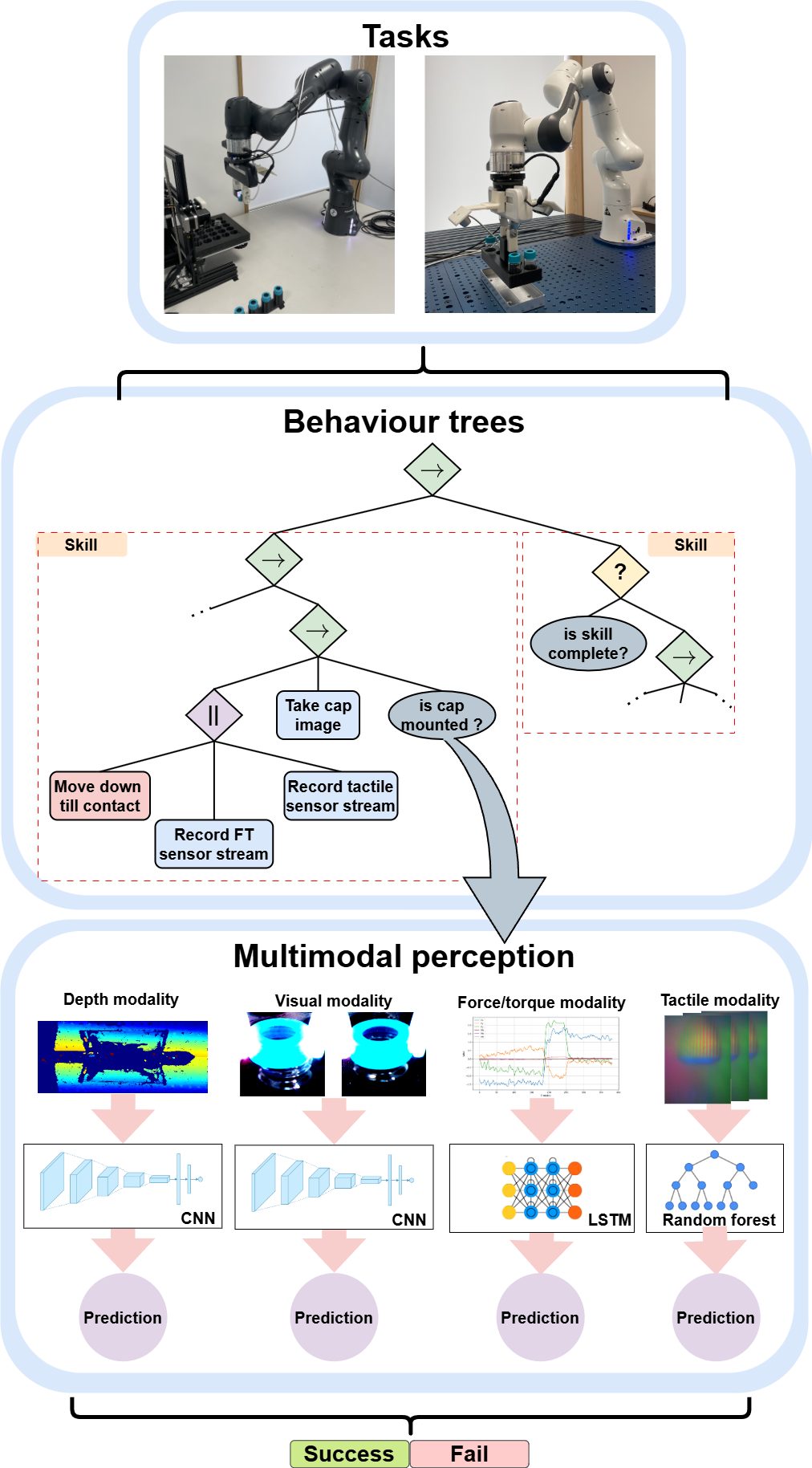}
    \caption{Overview of our proposed method. Lab tasks are represented using behaviour trees, which in turn used multimodal perception to assess the task execution status, thus improving its robustness and reliability}
    \label{fig:enter-label}
\end{figure}
\section{Introduction}
\label{section_introduction}
Accelerating materials discovery is fundamental in addressing global challenges such as renewable energy, sustainable materials development and healthcare. 
AI-driven, closed loop automated experiments will pave the way towards automating lab workflows where robots and autonomous systems can conduct long-term experiments. 
While there has been remarkable progress to introduce robotics in the field of laboratory automation~\cite{burger2020mobile},~\cite{Tom2024}, current methods of using sequential, pre-programmed tasks lack the adaptability needed for diverse laboratory environments.
Often, experiments involve handling hazardous materials, which pose risks to users and instruments in case of a failure. 
In addition, monitoring and usage of these systems are by non-expert robotic users (chemists) who may be unfamiliar with automated systems.
Adoption of these systems would greatly increase with a higher degree of explainability, even at the low-level tasks.

To date, autonomous laboratory experiments in materials discovery rely either on Finite State Machines (FSMs)~\cite{burger2020mobile},~\cite{Fakhruldeen2022} or more recently Large Language Models (LLMs)~\cite{Darvish2024}. 
Using FSMs and LLMs still faces challenges in complex robotics tasks, particularly in safety-critical environments.
To address current open gaps related to modularity, interpretability and adaptability to handle unforeseen changes or errors, we propose the use of Behaviour Trees (BTs)~\cite{Colledanchise2018} for autonomous materials discovery tasks in laboratory experiments. 
This approach allows us to break down complex tasks into simpler structures comprised of modular, reusable nodes or behaviours. 

Additionally, autonomous experimental workflows are most often executed in an open-loop fashion~\cite{burger2020mobile},~\cite{Fakhruldeen2022}.
This necessitates a high degree of engineering and are only successful in an empty, perfectly fixed environment without any failures, practically impossible to achieve in real-world laboratories.
We argue that we not only need adaptive, modular and explainable BTs, but these need to rely on data across multiple sensory modalities.
We believe that combining data across different modalities would lead to improved task success.

In this work we investigate the use of multimodal BTs across two different laboratory tasks: (i) vial capping, i.e. the process of picking up a vial cap, placing on the mouth of a vial and screwing it on the vial until sealed and (ii) rack insertion, i.e. picking up a rack and placing it into a rack holder. 
We model the two tasks using BTs, with decision nodes relying on the outputs from pre-trained models across different data modalities (vision, depth, haptic (Force/Torque (F/T)) and tactile).
We report the overall task success rate when deployed on real robotic chemists using laboratory glassware.
In summary, the contributions of this work are:
\begin{itemize}
    \item Modelling Chemistry laboratory tasks using behaviour trees towards increasing traceability, reactivity and adaptability;
    \item Using multimodal decision nodes in laboratory tasks, including visual, depth, haptic and/or tactile feedback significantly improved tasks' success rate metric compared to single-modalities; 
    \item A demonstration of the multimodal BTs in laboratory settings in two safety-critical, failure-prone tasks: sample vial capping and laboratory rack insertion.
\end{itemize}
    




\section{Related Work}
\label{section_related_work}

\subsection{Chemistry Lab Automation}
Recently, Chemistry lab automation has gained traction with the accelerated introduction of general-purpose robotic systems e.g. mobile manipulators~\cite{burger2020mobile,Zhu2024} and serial manipulators~\cite{Darvish2024, Lim2021,Shiri2021}. 
For e.g. the mobile robotic chemist~\cite{burger2020mobile} was first used for a photocatalysis experiment, and then combined with other robotic systems, repurposed for an autonomous solid-state workflow for powder x-ray diffraction~\cite{Lunt2024}.
However, whilst these are excellent endeavours in removing tedious and repetitive tasks from the scientists’ daily routines, the experimental workflows lack modularity due to their reliance on sequential state machines while also are heavily dependent on pre-programmed tasks.
In this work, we address this through concentrating our research endeavours are towards a more modular approach i.e. BTs.

\subsection{Behaviour Trees}

\begin{figure}[t]
    \centering
\includegraphics[width=0.49\textwidth]{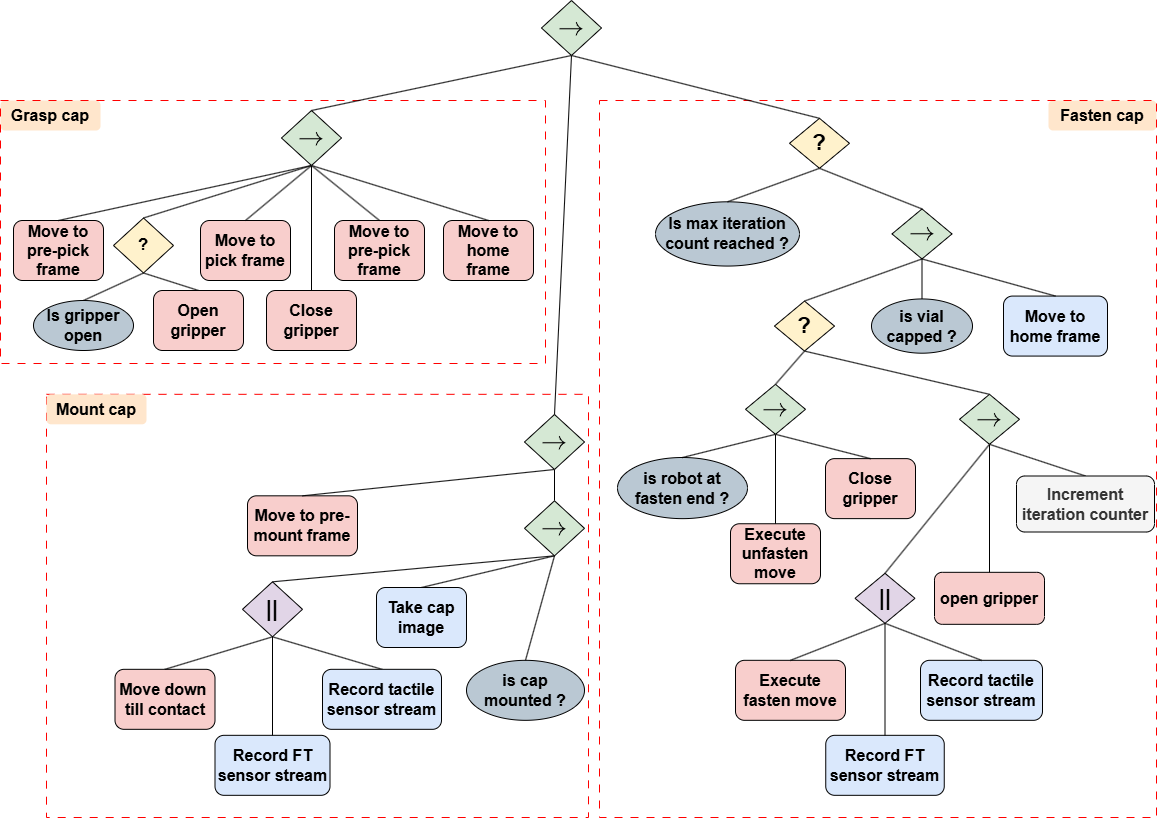}
    \caption{BT representation for the vial screw capping task. This task is composed of three skills: Grasp cap, mount cap and fasten cap}
    \label{fig:capping_bt}
\end{figure}

As BTs are a highly efficient way of creating complex systems that are both modular and reactive, they have been widely adopted in the robotics community across different tasks for e.g. multi-robots in multifarious environments~\cite{Best2024} and human-robot interaction~\cite{Scherf2023}.
It is therefore surprising that the natural sciences lab automation community still relies heavily on state machines for deploying the experimental workflows.
We aim to contribute to change this by demonstrating how BTs can be used for different tasks in autonomous Chemistry labs.

\subsection{Multimodal Sensory Information in Chemistry Lab Tasks}
To date, end-to-end autonomous materials discovery experiments have relied on open loop automation~\cite{burger2020mobile},~\cite{Lim2021},~\cite{Lunt2024} due to the high degree of reliability required for long-term experiments.
This has fuelled unprecedented levels of engineering and has made it non-trivial to transfer experimental setups between labs, resulting in increase the reproducibility challenges. 
Nonetheless, there have been efforts towards integrating different sensory modalities for single skills e.g. haptic feedback when scraping the wall of a vial~\cite{Pizzuto2024}, visual, F/T and tactile feedback when inserting vials in racks~\cite{Butterworth2023}, visual feedback for pouring liquids~\cite{Kennedy2019} and for high-accuracy needle injection in instrumentation~\cite{Angelopoulos2023}.
In these works, the authors focused on using perception methods for single skills to improve the task success.
Our work differs as we embed the BT decision nodes with the outputs from pre-trained models across a range of the four different data modalities.

\section{Behaviour Trees with Multimodal Sensory Information in Chemistry Lab Tasks}
\label{section_methodology}

\subsection{Behaviour Trees}
Behaviour trees are control structures used to represent plans and agent behaviours~\cite{Colledanchise2018}.
They are more human-readable and extensible than state machines or raw code due to their modular structure. 
In addition, they can be nested, enabling hierarchical representation across different abstraction level~\cite{Suddrey2022}.\par

A BT is a directed acyclic graph composed of nodes and edges. 
Each tree contains a single root node, with zero or more branches terminating at leaf nodes. 
Each node falls into into one of three categories: 1) leaf nodes that represent sensing and action primitives available to the agent 2) composite nodes that control branch selection, and 3) decorator nodes that modify branch outputs.\par
The execution of a BT starts at the root, ticking its children from left to right. 
Leaf nodes execute their goals and return their status directly, while composite nodes tick their children in a similar manner to root nodes and return their status based on their type and their children execution status.
Nodes return one of three statuses: running (still executing), success (goal achieved), or failure (goal not met). 
A special case of leaf nodes are \emph{condition} nodes that evaluate a proposition and only return success or failure.
Composite nodes include \emph{fallback}, \emph{sequence} and \emph{parallel} nodes. 
A fallback node returns success if any child succeeds and failure if all fail, while a sequence node fails if any child fails and only succeeds when all succeed. 
The parallel node ticks all children simultaneously and succeeds only when all children succeed. 
By combining these different node types, complex robotic tasks can be efficiently represented and programmed~\cite{Tanaka2023}.

\subsubsection{Task Disambiguation}
In our approach, tasks are represented as BTs. 
Each task is decomposed into skills, also modelled as BTs, which consist of primitive robotic actions such as movement and sensing commands. 
Movement commands involve positioning the end-effector or actuating the parallel grippers, while sensing commands capture data from available sensors. 
This data is then used by a multimodal condition node, described in the next section, to evaluate task execution. 
This structure simplifies complex tasks by breaking them into modular, reusable skills composed of basic robotic actions.

\begin{figure}[t]
    \centering    \includegraphics[width=0.3\textwidth]{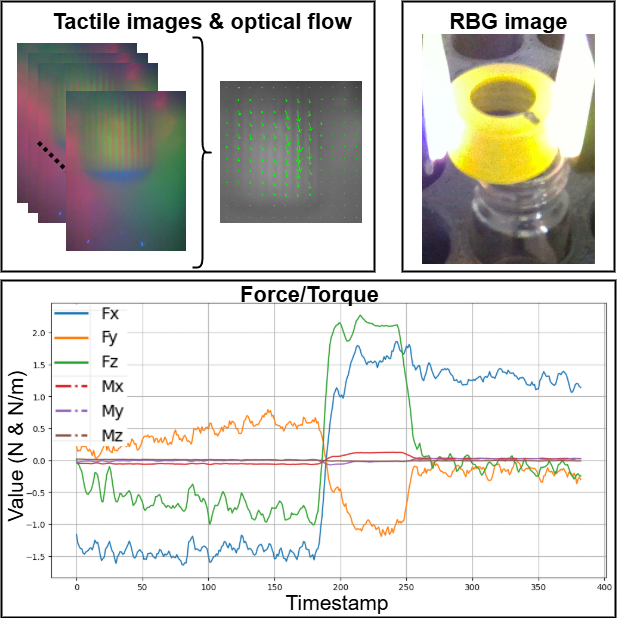}
    \caption{Example of data acquired from the vision, F/T and tactile modalities for the vial screw capping task.}
    \label{fig:capping_modalities}
\end{figure}

\subsubsection{Multimodal Condition Nodes}
To ensure each task phase is successfully completed prior to transitioning to the next one, condition nodes that utilise multimodal sensory data were added to the BT.
These nodes assess the current status by analysing data collected from previously executed robotic sensing commands. 
For each modality, a pre-trained model predicts whether the current phase succeeded or failed using the input data.

\vspace{-0.3cm}
\begin{equation}
\label{eq_v}
    is\_successful = \frac{\sum_{i=1}^{N}v_{i}s_{i}}{N} \geq \lambda
\end{equation}

The predictions are combined using the voting formula in Eq. \ref{eq_v}, which employs a weighted voting system across \( N \) modalities. Each modality has a manually assigned weight \( v_i \) (summing to one) that prioritizes its model's reliability and accuracy. Success (\( s_i = 1 \)) or failure (\( s_i = 0 \)) is weighted and averaged, and if the result meets or exceeds the threshold \( \lambda = 0.5 \), the task is deemed successful.  


\subsection{Multimodal Perception}
\label{sec:multimodal_perception}
Our multimodal condition nodes use different modalities based on their relevance to the specific task, selecting those most useful for capturing and assessing execution. Mainly four types of sensory data is utilised: RGB and depth from an RGB-D camera, haptic feedback from a wrist-mounted Force-Torque (F/T) sensor and tactile data from two fixed high-resolution tactile sensors on the parallel gripper.  
The heterogeneous nature of this data requires domain-specific models to capture the unique characteristics of each modality.

For visual and depth feedback, we use convolutional neural networks (CNNs)-based architectures e.g. Residual Networks (ResNet)~\cite{He2015} and the VGG-19 network~\cite{Simonyan2014}. 
We feed images directly from the sequence recorded through the camera on the robot's end-effector, attached via a 3D printed mount. 
This mount was designed to capture task-specific images, maximising information to help the model make accurate decisions. 
All images were recorded using a framerate of 30 frames per second.

For F/T feedback, a Bidirectional Long Short-Term Memory (Bi-LSTM) architecture was used~\cite{Hochreiter1997}. 
While LSTM networks are a type of recurrent neural network that excel at capturing dependencies in sequential data by maintaining information over time, Bi-LSTM processes sequential data in both forward and backward directions. From the six-axis F/T sensor, N readings are recorded and fed to the model as a $N\times6$ time series, where N is task specific and depends on the action duration. Given the noisiness of the F/T data and its susceptibility for sensor drift, input data was zero-centred and smoothed using a moving average filter.


For the tactile modality, the input sequence of tactile images is used to extract a specific range that captures the time window during which changes in tactile data occur due to contact.
Optical flow, which measures the motion of objects or surfaces between consecutive frames in a video or image sequence, was used to compute the maximum flow between frames \cite{baker2004lucas}. 
We then extracted features such as entropy, energy, and mean magnitude from the optical flow data. 
These features were used to train a random forest classifier, which combines the outputs of an ensemble of decision trees to improve accuracy and make a reliable prediction for the given task~\cite{Ho1995}. 





\section{Experimental Evaluation}
\label{section_experimental_evaluation}

Our experiments evaluate the effectiveness of our proposed method for automating lab tasks, focusing on reliability and robustness. Specifically, we investigate: 1) BT representation suitability for chemistry lab automation; 2) Performance of available sensing modalities; 3) Performance improvements using multimodal perception versus individual modalities.\par
To address these questions, we applied our methodology to two common chemistry lab tasks.
First, we examine sample vial screw capping, essential for sample integrity during storage and transport. Using vials identical to those in prior autonomous solid-state powder x-ray diffraction work~\cite{Lunt2024} (previously achieved with a dual-arm robot without sensory feedback), we aim to accomplish this with a single arm, minimizing potential human exposure during vial transport.
Second, we consider laboratory rack insertion: a manipulator inserts a vial rack into a holder. This task is common in chemistry workflows for rack transport between instruments~\cite{burger2020mobile},~\cite{Fakhruldeen2022}.

\subsection{Robotic Setup}
The robot for both tasks is the Panda Franka-Emika 7 degrees-of-freedom manipulator with a Franka hand gripper. \par 
For the vial screw capping task, the setup had the following additional sensors: a Robotiq FT300 sensor for measuring F/T at the robot's wrist along three axes, an Intel RealSense D435i RGB-D camera mounted on the gripper and two DIGIT tactile sensors attached to the gripper fingers~\cite{digit2020}. 
A liquid handling gantry system for 20ml vials and a five-slot cap holder were positioned within the robot's workspace.\par

For the rack insertion task, a similar setup was used, with an additional D435i camera mounted on the opposite side of the gripper to cover both sides of the rack during insertion. No tactile sensors were deemed necessary for this task. 
A custom-made 6-vial rack, based on the design from \cite{burger2020mobile}, was employed. 
Both setups are shown in Figure~\ref{fig:enter-label}.

\subsection{Laboratory Task I: Sample Vial Screw Capping}

\subsubsection{Task Formulation Using BTs} 
\label{sec:capping_modelling}

This task was represented using the BT illustrated in Figure \ref{fig:capping_bt}. It is comprised of a sequence node that goes executes the skills: \emph{grasp cap} where the robot picks up a cap from a cap holder; \emph{mount cap} where the robot mounts the held cap on top of the vial neck; and \emph{fasten cap} where the robot securely affix the cap to the vial.\par

The \emph{grasp cap} skill is represented by a sequence node that first commands the robot to move to the cap pre-pick frame. Next, a fallback node checks if the gripper is open; if not, the gripper is opened. Next, the following robot actions are executed sequentially: move to the cap pick position, close the gripper, return to the pre-pick position, and then move to the home position. The \emph{mount cap} skill is also a sequence node. It starts with a command to move to the pre-mount position, followed by another sequence node. This sequence first ticks a parallel node that simultaneously executes three actions: move downwards until contact is detected, record F/T feedback for 2 seconds, and capture tactile data for the same duration. Once these actions succeed, the robot captures an RGB image and then using a multimodal condition node checks if the cap is properly aligned. If misaligned, the cap screw threads won't mesh with the vial's ones and it will fall when the gripper releases it.\par

The \emph{fasten cap} skill is controlled by a fallback node that first checks if the maximum iteration count is reached, preventing infinite retries if the multimodal condition fails. If not, a sequence node is ticked. This sequence starts with a fallback node that alternates between fastening the cap while recording data and resetting the robot for the next attempt. The first child of this fallback node checks if the robot is at the end of the fastening move. If true, the next action moves the robot gripper back to the starting position (a 90$^o$ counterclockwise rotation). Then, the gripper closes, ending this sequence. If the robot is already at the start position, this node fails, and the next sequence node is ticked. In the next sequence, a parallel node is ticked. It simultaneously fastens the cap (a 90° clockwise rotation) while recording force/torque and tactile data for 4 seconds. After the parallel node succeeds, the gripper open, and a leaf node increments the fasten iteration counter, concluding the fallback node's execution. Finally, the multimodal condition node checks if the vial is fully capped which means the cap is securely attached to the vial and no more fastening is needed. If successful, the robot moves to the home frame. If not, the top fallback node is triggered again, repeating the process.

\subsubsection{Perception Modalities}
\label{sec:capping_modalities}

In the \emph{mount cap} skill, vision, F/T and tactile modalities were used. For each modality, a model was trained to classify if the cap was properly or improperly mounted on the vial. Figure~\ref{fig:capping_modalities} shows examples of the data acquired for the modalities used in this task.

For the visual modality, the ResNet-18 architecture was used. 
Starting with the pre-trained Imagenet weights, the model was finetuned to adapt to our dataset that included $600$ images per class for a total of two classes. 
Accordingly, the final fully connected layer was adjusted to output two classes for binary classification. 
Early stopping, learning rate schedulers and data augmentations were applied to enhance generalisation when training. 
Our trained model achieved testing accuracy of $98.8\%$, as shown in Table~\ref{tab:model_performance_capping}. To test the model performance on the real system, $50$ trials were conducted to evaluate our model, alternating between proper and improper mounting scenarios. For the improper mount cases, errors were introduced by means of small random offsets (in the range of $\pm10$ mm, which is equal to the vial's neck radius) applied to the end-effector $x$ and $y$ translation axes at the pre-mount frame. This offset simulates errors from misplacing the vial or cap. The model classification accuracy dropped to $88\%$. This can be attributed to the challenges of distinguishing between classes when the cap is very close to the correct position, compounded by the relatively small dataset used for training the model.\par

To train the F/T model, $300$ sequences per class were used, where each individual time stamp was labelled and fed into the model when training. The used architecture followed a conventional Bi-LSTM setup, with dimensions of (6,50,1) for the input, hidden and output layers respectively. A dropout was added to prevent over-fitting. Predictions were aggregated by averaging the outputs from each time stamp and applying a $0.5$ threshold for final classification. The model achieved a test accuracy of $95\%$ and a deployment accuracy of $96\%$ when tested for $50$ trials as before. For the tactile modality, a random forest classifier was trained with $300$ trials per class using the method described in section~\ref{sec:multimodal_perception}. To improve the model prediction accuracy, hyper parameter optimisation was carried out. This model achieved an $85.9\%$ test accuracy and $82\%$ deployment accuracy. The lower model accuracy is due to the similar tactile deformation patterns observed in both cases, which makes it challenging for the model to accurately classify sequences when the cap is near the correct position.\par 

For the \emph{fasten cap} skill, both F/T and tactile modalities were utilised. Using the same approaches as described for the previous skill, new models were trained to detect when the vial was fully capped, meaning the screw cap was completely tightened, ensuring the vial was securely sealed. The vision modality was excluded, as the images produced by the camera setup did not provide clear distinctions between classes that could not even be identified by human observers. A dataset of 200 trials per class was used to train these models. The test accuracy achieved was 99\% for the F/T model and 98\% for the tactile model. To measure deployment accuracy, 50 trials were conducted for each modality, alternating between successful and unsuccessful vial capping. In the successful case, the cap was positioned near the fully capped state, allowing it to lock during the fastening motion. In the unsuccessful case, the cap was left in a position where it would not lock at any point during the movement. Deployment accuracy was 94\% for the FT model and 76\% for the tactile model. The comparatively lower accuracy of the tactile model is due to the incorrect prediction of unsuccessful capping as successful. This is caused by several factors, including cross-threading or contamination of the threads. However, this presents an ideal case to demonstrate the effectiveness of our multi-modal perception approach. \par

\vspace{-1em}
\begin{table}[h]
\caption{Overview of model performance for different modalities for the vial screw capping task; both when training the models and deploying them on the real system}
\label{tab:model_performance_capping}
\centering
\begin{tabular}{l|llcc}
\textbf{Skill}             & \textbf{Modality} & \textbf{Model} & \textbf{\begin{tabular}[c]{@{}c@{}}Test\\ Accuracy\end{tabular}} & \textbf{\begin{tabular}[c]{@{}c@{}}Deployment\\ Accuracy\end{tabular}} \\ \hline
\multirow{3}{*}{Mount cap}  & Vision            & Resnet18       & 98.8\%                                                           & 88\% $\pm$ 2.3                                                            \\
                           & Force/Torque      & Bi-LSTM         & 95\%                                                             & 96\% $\pm$ 1.39                                                           \\
                           & Tactile           & RandomForest   & 85.9\%                                                           & 82\% $\pm$ 2.72                                                           \\ \hline
\multirow{2}{*}{Fasten cap} & Force/Torque      & Bi-LSTM         & 99\%                                                             & 94\% $\pm$ 1.68                                                           \\
                           & Tactile           & RandomForest   & 98.3\%                                                             & 76\% $\pm$ 3.02                                                          
\end{tabular}
\end{table}


\begin{figure}[t]
    \centering
    \includegraphics[width=0.45\textwidth]{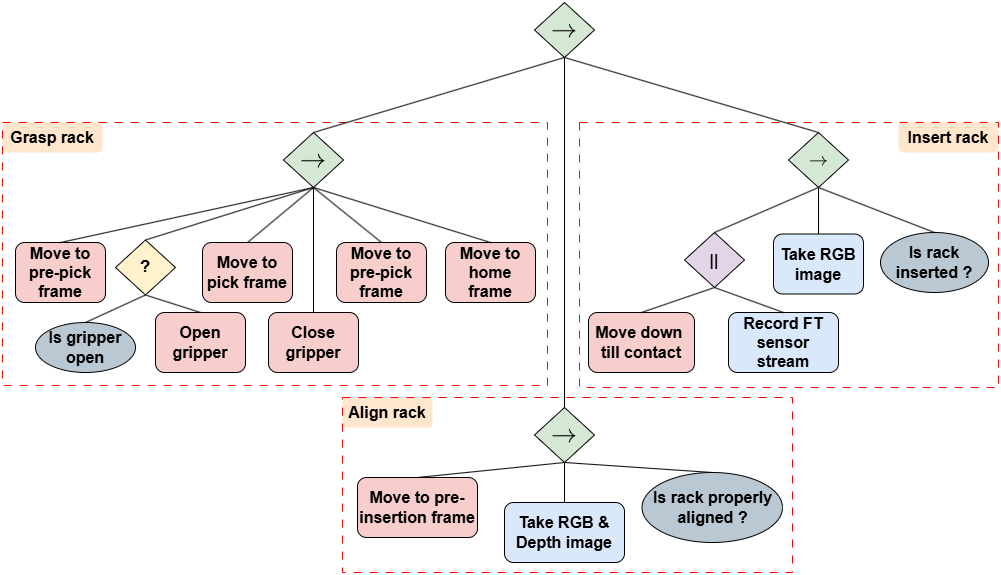}
    \caption{The BT representation for the rack insertion task. This task is composed of three skills: grasp rack, align rack and insert rack.}
    \label{fig:rack_insertion_bt}
\end{figure}

\subsubsection{Validation of Single Skills}
The \emph{mount cap} skill was tested by conducting 50 trials, alternating between proper and improper mounting
scenarios similar to the process described in section \ref{sec:capping_modalities}. The average execution duration was 36.7 seconds with a success rate of 94\%. Similarly, the fastening phase was tested by conducting 50 trials, with trials alternating between successful and unsuccessful vial capping. The average execution time was 33.5 seconds with success rate of 100\%.  These results are listed in table \ref{tab:capping_individual_results}. By comparing the accuracy data in table \ref{tab:model_performance_capping} with the success rates in table \ref{tab:capping_individual_results}, it is evident that combining multiple modalities yields better results. The multi-modal condition node ensures that the prediction error of one model is compensated by the high accuracy of another by assigning correct weights and threshold. For example, although the fastening attains 100\% (table \ref{tab:capping_individual_results}), the tactile model made prediction errors 13 out of 50 times. However, the 98\% accuracy of the FT model compensated for these errors. Also, it can be observed from the tables \ref{tab:model_performance_capping} and \ref{tab:capping_individual_results} that, by combining the modalities, average success rates of vial capping task improved by 11.7\% (6.09\% in the mounting cap skill and 17.65\% in the fasten cap skill).

\begin{table}[h]
\caption{Results of testing the individual phases for vial capping task}
\label{tab:capping_individual_results}
\centering
\begin{tabular}{lccc}
\textbf{Skill} & \textbf{\begin{tabular}[c]{@{}c@{}}Number of \\ Runs\end{tabular}} & \textbf{\begin{tabular}[c]{@{}c@{}}Avg. duration\\ (secs)\end{tabular}} & \multicolumn{1}{l}{\textbf{Success rate}} \\ \hline
Mount cap       & 50                                                                 & 36.73                                                                    & 94\% $\pm$ 1.68                              \\
Fasten cap      & 50                                                                 & 33.47                                                                    & 100\%                           
\end{tabular}
\end{table}

\subsubsection{End-to-end Task Evaluation}
To test the complete task, 25 trials were conducted with the success criteria being the vial securely capped by the end of the task without exceeding the max iteration count for the fastening phase. The maximum count was determined manually by allowing the robot to perform the full fastening operation multiple times, each time starting from a different position. As shown in table \ref{tab:asks_results}, the achieved success rate was 88\% with an average execution time of 327.72 seconds. One failure occurred because the maximum iterations were exceeded before sealing the vial. The other two were due to the BT incorrectly concluding success before the vial was fully sealed.

\begin{table}[h]
\caption{Results of testing end-to-end tasks}
\label{tab:asks_results}
\centering
\begin{tabular}{lccc}
\textbf{Task} & \textbf{\begin{tabular}[c]{@{}c@{}}Number of \\ Runs\end{tabular}} & \textbf{\begin{tabular}[c]{@{}c@{}}Avg. duration\\ (secs)\end{tabular}} & \multicolumn{1}{l}{\textbf{Success rate}} \\ \hline
Rack insertion       & 25                                                                 & 3.73                                                                    & 92\% $\pm$ 1.92                               \\
Vial screw capping      & 25                                                                 & 327.72                                                                    & 88\% $\pm$ 1.62                              
\end{tabular}
\end{table}


    

    

\subsection{Laboratory Task II: Rack Insertion}
\subsubsection{Task Formulation Using BTs}
\label{sec:insertion_modelling}
The BT representation for this task is shown in figure \ref{fig:rack_insertion_bt}. This task is represented by a sequential node that ticks through the following skills: \emph{Grasp rack} where the robot picks up a rack from the table; \emph{align rack} where the robot aligns the held rack with its holder; and \emph{insert rack} where the robot inserts the rack into the holder.\par

The skill \emph{grasp rack} reuses the \emph{grasp cap} skill described in section \ref{sec:capping_modelling}, highlighting the modularity of using BTs. The \emph{align rack} skill is represented by a sequence node that ticks through the following actions: moving the robot to the pre-insertion position, acquiring RGB and depth images from the cameras, and a multimodal condition node that checks rack alignment using the images. If the rack is misaligned, the insertion will fail. The \emph{insert rack} skill is also a sequence node. It begins with a parallel node that simultaneously moves the robot downward until contact is detected and records F/T feedback for 2 seconds. Once successful, an RGB image is captured, followed by a multimodal condition node to confirm if the rack was properly installed in the holder.



\begin{figure}[h]
    \centering
    \includegraphics[width=0.37\textwidth]{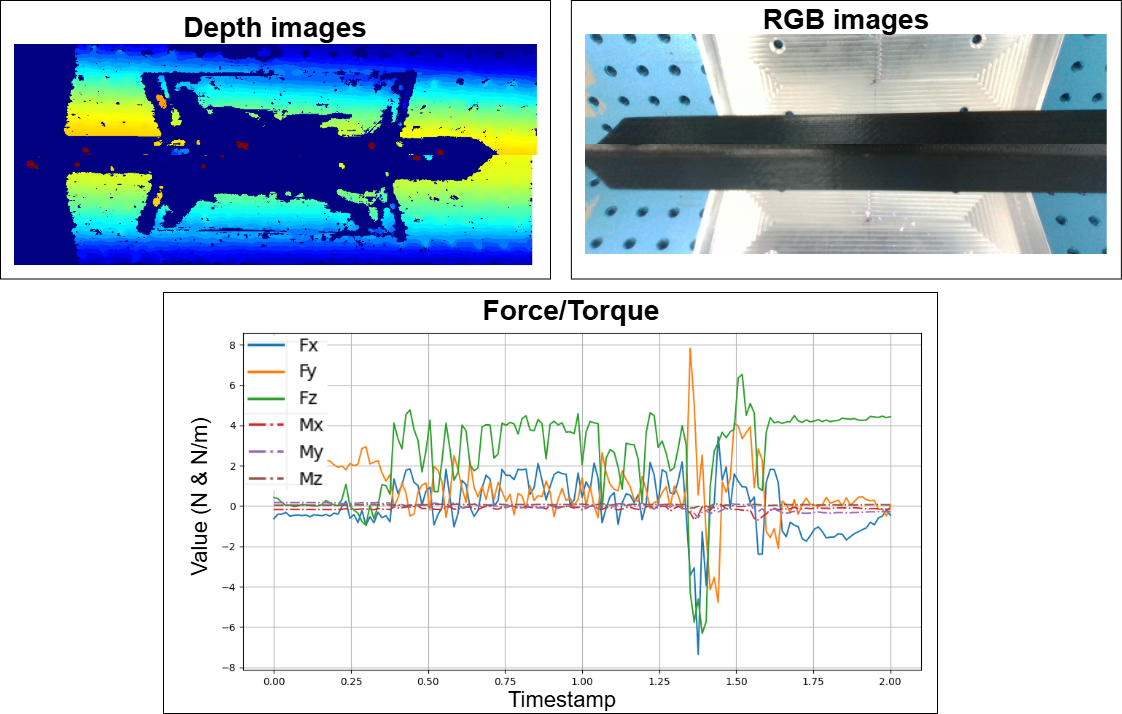}
    \caption{Example of data acquired from the vision, depth and F/T modalities for the rack insertion task.}
    \label{fig:insertion_modalities}
\end{figure}

\subsubsection{Perception Modalities}
\label{sec:insertion_modalities}
 Vision and depth modalities were used for evaluating the \emph{align rack} skill. 
 For each modality, a model was trained to predict if the rack was aligned with the rack holder. Figure~\ref{fig:insertion_modalities} illustrates the data acquired for the modalities used in this task.

For each modality, a VGG-19 network was trained using feature extraction, starting from a pre-trained model on the ImageNet dataset. 
Since two cameras were used, features from each camera were extracted using separate VGG-19 networks, then concatenated and passed through a fully connected network. A sigmoid activation function was applied after the fully connected layer to produce a probability score for binary classification, with binary cross-entropy used as the loss function during training. The training dataset comprised of $3500$ images per class, followed by data augmentation methods. The test accuracy for the vision and depth modalities were $98\%$ and $99\%$ respectively (Table~\ref{tab:insertion_modalities}).\par

To evaluate model deployment on the robot, $50$ trials were conducted, alternating between scenarios where the rack was properly aligned with the holder and those where alignment errors were introduced by means of small random offsets applied to the end-effector $x$ and $y$ translation axes and rotation around the z-axis at the alignment frame. 
The vision model's deployment accuracy was $95\%$ and the depth model's deployment accuracy was $80\%$.

For the \emph{insert rack} skill, vision and F/T modalities were used. 
Depth data was excluded due to poor image quality caused by proximity to the table surface. Similar to the previous skill, a VGG-19 network was trained with a dataset of $600$ images per class to classify successful rack insertion. Model test accuracy was $99\%$. To train the model for the F/T modality, $600$ sequences per class were used, using the same Bi-LSTM architecture and approach as described in section~\ref{sec:capping_modalities}. The model achieved a test accuracy of $95\%$.\par

The model's performance was evaluated over $50$ trials, alternating between successful and failed rack insertions. 
Failures were induced by applying small random offsets to the robot's end-effector, preventing the rack from entering its holder. The resulting deployment accuracy were $94\%$ and $90\%$ for the F/T and vision models respectively. 

\vspace{-1em}
\begin{table}[h]
\caption{Overview of model performance for different modalities for the rack insertion task; both when training the models and deploying them on the real system}
\label{tab:insertion_modalities}
\centering
\begin{tabular}{l|llcc}
\multicolumn{1}{c|}{\textbf{Skill}} & \multicolumn{1}{c}{\textbf{Modality}}                              & \multicolumn{1}{c}{\textbf{Model}} & \textbf{\begin{tabular}[c]{@{}c@{}}Test\\ Accuracy\end{tabular}}               & \textbf{\begin{tabular}[c]{@{}c@{}}Deployment\\ Accuracy\end{tabular}}         \\ \hline
\multirow{2}{*}{Align rack}          & Vision                                                       & VGG-19                             & 98\%                                                                           & 95\% $\pm$ 1.54                                                                           \\
                                    & Depth                                                             & VGG-19                                & 99\%                                                                           & 80\% $\pm$ 2.83 \\ \hline
\multirow{2}{*}{Insert rack}          & Force/Torque                                                       & BiLSTM                             & 99\%                                                                           & 94\% $\pm$ 1.68                                                                           \\
                                    & Vision                                                             & VGG-19                                & 99\%                                                                           & 90\% $\pm$ 2.12                                                                          
\end{tabular}
\end{table}

\subsubsection{Validation of Single Skills}
To test the alignment skill, 50 trials were conducted, alternating between properly aligning the rack and not. The average execution duration was $2.55$ seconds with a success rate of $95\%$. Similarly, the insertion skill was tested by conducting 50 trials, with trials alternating between successful and failed insertion. The average execution time was $1.2$ seconds with success rate 0f $91\%$. These results are listed in table \ref{tab:insertion_individual_results}. A $3.91\%$ improvement in the average success rate of the rack insertion task can be observed from Tables \ref{tab:insertion_modalities} and \ref{tab:insertion_individual_results}.

\vspace{-1em}
\begin{table}[h]
\caption{Results of testing the individual phases for rack insertion task}
\label{tab:insertion_individual_results}
\centering
\begin{tabular}{lccc}
\textbf{Skill} & \textbf{\begin{tabular}[c]{@{}c@{}}Number of \\ Runs\end{tabular}} & \textbf{\begin{tabular}[c]{@{}c@{}}Avg. duration\\ (secs)\end{tabular}} & \multicolumn{1}{l}{\textbf{Success rate}} \\ \hline
Align rack       & 50                                                                 & 2.55                                                                   & 95\% $\pm$ 1.54                               \\
Insert rack      & 50                                                                 & 1.2                                                                    & 91\% $\pm$ 2.02                              
\end{tabular}
\end{table}

\subsubsection{End-to-end Task Evaluation}
To test the complete task, $25$ trials were conducted with the success criteria being the rack being correctly inserted at the end of execution. As shown in table \ref{tab:asks_results}, the achieved success rate was $92\%$ with an average execution time of $3.73$ seconds.

\section{Conclusion}

In this work, we demonstrated that multimodal BTs 
are a viable approach towards making closed-loop robotics lab task automation modular, adaptive and interpretable. 
We explored four different sensory modalities: visual, haptic (F/T), tactile and depth across two safety-critical, failure-prone Chemistry lab tasks.
Our empirical results demonstrate that our proposed method enables the robot to complete the tasks successfully with over $88\%$ task success across numerous trials.
There are several future directions for extending and improving this approach. 
On our immediate future plan, we want to explore the addition of recovery actions within the BT and to carry out user studies to understand better how chemists would find the benefits of the multimodal BTs as compared to previous, more rigid open-loop methods based on state machines.

\bibliographystyle{ieeetr}
\bibliography{references}

\end{document}